# Palmprint Recognition in Uncontrolled and Uncooperative Environment

Wojciech Michal Matkowski, Tingting Chai and Adams Wai Kin Kong

*Abstract*— Online palmprint recognition and latent palmprint identification are two branches of palmprint studies. The former uses middle-resolution images collected by a digital camera in a well-controlled or contact-based environment with user cooperation for commercial applications and the latter uses high-resolution latent palmprints collected in crime scenes for forensic investigation. However, these two branches do not cover some palmprint images which have the potential for forensic investigation. Due to the prevalence of smartphone and consumer camera, more evidence is in the form of digital images taken in uncontrolled and uncooperative environment, e.g., child pornographic images and terrorist images, where the criminals commonly hide or cover their face. However, their palms can be observable. To study palmprint identification on images collected in uncontrolled and uncooperative environment, a new palmprint database is established and an end-to-end deep learning algorithm is proposed. The new database named NTU Palmprints from the Internet (NTU-PI-v1) contains 7881 images from 2035 palms collected from the Internet. The proposed algorithm consists of an alignment network and a feature extraction network and is end-to-end trainable. The proposed algorithm is compared with the state-of-the-art online palmprint recognition methods and evaluated on three public contactless palmprint databases, IITD, CASIA, and PolyU and two new databases, NTU-PI-v1 and NTU contactless palmprint database. The experimental results showed that the proposed algorithm outperforms the existing palmprint recognition methods.

## I. INTRODUCTION

A number of biometric characteristics such as face, fingerprint, palmprint, iris, gait, voice, and handwriting have been proposed. Some of them, e.g., fingerprint and iris have already achieved very high accuracy [1] and been commercially deployed. Many face recognition methods are already close to human-level performance [2], [3]. Law enforcement agencies have been using fingerprints for searching suspects from the early 20th century [4]. Iris, voice and palmprint recognition also perform very well [1]. Each recognition system is designed to operate on the traits acquired under a specific environment. In a constrained environment, control over some of data acquisition parameters is assumed whereas in an uncontrolled and uncooperative environment, there is no such assumption. Although several biometric areas are very successful and various research studies have been done, the recognition in the uncontrolled and uncooperative environment is still challenging and some forensic applications are not well investigated.

An important part of forensic investigation is criminal and victim identification based on evidence images. The identification from evidence images is very troublesome if no obvious traits such as face or tattoos are available. Terrorists, rioters, child sexual offenders usually cover or hide their faces or tattoos to avoid identification, but other body parts, e.g., hands, forearms, back, chest, and legs can be still visible. Sometimes, criminals also try to hide the identity of their victims. The cases without face and tattoos have been considered in some recent studies to develop biometric traits such as vein [5], skin mark [6], androgenic hair [7] and skin texture [8] for forensic applications. Veins and skin marks require high-resolution image [9]. Some of the body parts have no or not enough skin marks or hair and skin texture may not be discriminative enough if only small skin region is available [8]. Still, open hands can be observed in some of the images [10], [11], particularly when the subjects salute, wave, raise their hands, touch the victim or the offender, or try to cover the camera. These evidence images can be collected from the Internet, social media, the suspect's hard drives, suspect's or victim's families or during police bookings, etc., and can be used for forensic investigation. Fig. 1 shows some potential evidence images collected from the Internet, which include subjects exposing their palms.

Palm features, including flexion creases, wrinkles, ridges, and minutiae are located on a palmar side of a hand and are considered to be permanent and unique to an individual [12]. Palmprint has been studied for over two decades. The previous studies can be classified into contact-based, contactless, latent [12] and 3-D palmprint [13] (see Fig. 2). Many methods have been proposed, achieving high recognition performance, especially on publicly available palmprint databases,

Manuscript received [date]. This work is partially supported by the Ministry of Education, Singapore through Academic Research Fund Tier 2, MOE2016-T2-1-042(S).

W.M. Matkowski is with the School of Computer Science and Engineering, Nanyang Technological University, Singapore, 639798. E-mail: matk0001@e.ntu.edu.sg.

*T. Chai is with the Institute of Information Science, Beijing Jiaotong University, Beijing, 100044, P.R. China. E-mail: ttchai@bjtu.edu.cn.

A.W.K. Kong is with the School of Computer Science and Engineering, Nanyang Technological University, Singapore, 639798. E-mail: adamskong@ntu.edu.sg.

*All her work was done in the Nanyang Technological University, Singapore.

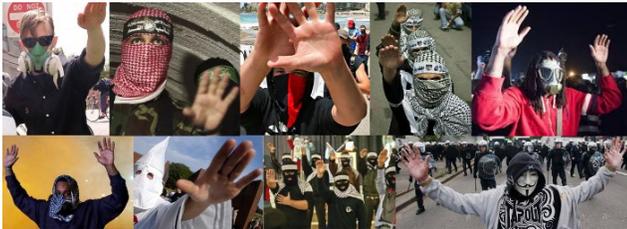

Fig. 1. Examples of images containing terrorists and rioters exposing their palms. Child sexual abuse images also contain hands; however they cannot be put here due to legal reasons.

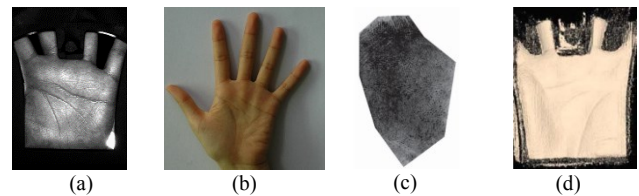

Fig. 2. Different types of palmprints studied by the biometric community. (a) Contact-based [12], (b) contactless, (c) latent [24], and (d) 3D [13].

established in environments, where image acquisition parameters such as illumination, background, type of camera, etc. can be controlled [13]. The typical pipeline of palmprint recognition consists of a sequence of modules responsible for pre-processing, segmentation, palm region of interest (ROI) extraction, feature extraction and matching. The modules are designed to extract desired features at each stage and forward the useful information to the next module. To successfully process the information, in most of the existing methods, the features are carefully hand-crafted using human knowledge about the hand, palm and also the image acquisition settings.

A recent success of deep learning, including convolutional neural networks (CNN) [14] in computer vision, has been encouraging researchers to shift from the traditional feature-engineering approach to the deep learning approach. In the deep learning approach, rather than designing features in each module separately, the network architecture is designed and during the network training, the hierarchical data representations are being learned, typically using backpropagation with stochastic gradient descent (SGD) optimization [15]. The success of deep learning highly depends on the amount of training data. In the image based biometrics, the shift occurred especially in the face recognition domain where large datasets are already available and many deep learning methods [2], [3], [14], [17] outperformed traditional face recognition methods on very challenging unconstrained benchmarks such as Labelled Faces in the Wild (LFW) or YouTube Faces (YTF) with significant margins. In addition to face, CNNs were also applied to fingerprint [18], iris [19] and palmprint [13], [20], [21], [22], [23] recognition.

The previous palmprint research focused mostly on online palmprint recognition for commercial applications and latent palmprint identification in crimes such as robbery and homicide. The community neglected the uncontrolled and uncooperative palmprint recognition problem mentioned before and the merit of palmprint recognition for forensic investigation based on digital images is not fully exposed yet. Contrary to the online palmprint recognition, the uncontrolled and uncooperative palmprint recognition problem has different requirements, especially in terms of the imaging environment and the recognition algorithm has to be applicable to low-resolution images taken under uncontrolled and contactless environment without subject's cooperation. Comparing with latent palmprint recognition, the problem studied in this paper has no high-resolution features, e.g., minutia and ridges. Most of the previous studies use the feature-engineering approach and the application of deep learning to palmprint recognition received moderate attention from the biometrics community.

This paper aims to address the uncontrolled and uncooperative palmprint recognition problem by establishing a new palmprint database named NTU Palmprints from the Internet database (NTU-PI-v1) containing 7881 images from 2035 palms collected from the Internet and proposing an end-to-end deep learning algorithm composed of an alignment network, a feature extraction network and an in-network data augmentation scheme. The rest of this paper is organized as follows. In Section II, related works including palmprint recognition methods and contactless palmprint benchmarks: CASIA, IITD and PolyU are described. In Section III, the NTU-PI-v1, the largest palmprint database, is presented. In Section IV, the proposed network architecture design is provided. In Section V, the evaluation protocols, implementation details, training schemes, experimental results and comparison with the state-of-the-art contactless palmprint recognition methods on five databases, CASIA, IITD, PolyU, NTU contactless palmprint database (NTU-CP-v1) and NTU-PI-v1 are reported. In Section VI, the conclusions and discussion are given.

## II. RELATED WORK

The related conventional and some recent deep learning based palmprint recognition methods are first described. Then, the contactless palmprint benchmark databases, including CASIA, IITD and PolyU, which are commonly used for palmprint recognition evaluation, are presented.

### A. Palmprint Recognition Methods

So far, no one studies palmprint recognition in an uncontrolled and uncooperative environment. The most related one is contactless palmprint recognition, which has been addressed by a number of researchers [25], [26], [27]. In a contactless setting, the image is taken by a camera without physical contact between the hand and any surface of the acquisition device. This setting usually results in some rotation, translation, scale and illumination variations, depending on the acquisition device. The acquired image, consisting of a full or partial palmar side of a hand is input and processed by a recognition method. The goal of pre-processing and segmentation is to extract the hand contour and localize the hand landmarks, usually situated between fingers. In a well-controlled environment where the background has single color, Gaussian smoothing, thresholding, edge detectors and boundary tracking methods give satisfactory pre-processing and segmentation results. Then, to align palmprints into the same coordinate system, the landmarks are used to define the location of the palmprint ROI [12], [28]. Originally, to capture and extract the palmprint features, coding based methods such as PalmCode [28], CompCode [29], and OrdinalCode [30] were proposed for the contact-based environment. However, according to [13], they usually do not perform well on contactless palmprints, mostly because of image misalignment. Thus, some studies suggested using more robust features, which employ local descriptors such as Scale Invariant Feature Transform (SIFT) [25], Local Binary Pattern (LBP) based features [31], [26], local micro-structure tetra pattern [32], local line directional pattern (LLDP) [33] and histogram of oriented lines (HOL) [34]. Recently, several new coding based methods such as Double Orientation Code (DOC) [35] and Difference of Normals (DON) [36] which extracts 3D information from a 2D image were also proposed. Moreover, some studies proposed learning based descriptors such as CR-CompCode [37] and discriminant direction binary code (DDBC) [38] to learn palmprint features.

To the best of our knowledge, there are few systematic studies on deep learning application to palmprint recognition. Svoboda et al. [20] proposed to use a Siamese network, with an architecture similar to AlexNet [14], for palmprint recognition. During training, the network is fed with triplets of palmprint ROIs (128 by 128 pixels) from different images and outputs 32-dimensional feature vectors for calculating the $d$-prime loss. This method requires a predefined triplet generation scheme and at least two samples collected from each subject's palmprint because each triplet contains two different samples coming from the same palm and one sample from a different palm. During the recognition, the comparison scores are determined by the distances between the feature vectors of two input samples. Svoboda et al. evaluated the proposed architecture on two contactless palmprint benchmarks and compared it with OrdinalCode and CompCode. Minaee et al. [39] employed a deep scattering convolutional network (DSCN) [40] for feature extraction with principal component analysis (PCA) and support vector machine (SVM) for recognition, achieving very good results. However, the network consists of fixed, non-trainable filters and the evaluation was performed on a contact-based dataset. Meraoumia et al. [41] employed a PCANet [42] for multispectral palmprint recognition. However, the authors did not perform experiments on contactless palmprint databases. In the most recent palmprint recognition survey [13], Fei et al. evaluated a number of feature extraction methods for contactless palmprint recognition, including four deep learning architectures, AlexNet, VGG-16, GoogLeNet and Res-Net-50. These networks were pre-trained using the ImageNet dataset and then palmprint ROI images were used to fine-tune the networks. The authors concluded that deep learning methods achieve comparable or even higher performance than conventional palmprint methods. Ramachandra et al. [21], investigated contactless palmprint recognition of newborn babies using a pre-trained AlexNet. The authors manually cropped ROI to fine-tune AlexNet and trained SVM using the features from the last fully connected layer. For recognition, the sum rule of SVM and softmax scores was used. Genovese et al. [22], proposed a 3-layer network called

PalmNet, which combines pre-defined Gabor filters in the first layer, PCA in the second layer and binarization in the third layer. The authors also proposed a training algorithm for PalmNet, which employs Gabor filter selection and PCA. The network was trained using ROIs extracted by a traditional method and evaluated on four contactless palmprint databases, achieving good results. Other works [43], [44], [45] performed a rather ambiguous evaluation. Dian et al. [43] used AlexNet for extracting features from palm ROIs and Hausdorff distance for matching but they did not compare with other palmprint recognition methods. Jalali et al. [44] tried to address deformation invariance by using a small CNN but the evaluation was only done on a very small dataset of 200 palmprint images (28x28 pixels) from 10 persons only and the CNN was trained using 85% of the dataset. Bao et al. [45] used CNN for landmark detection and the same method as in [28] to define the ROI based on two detected landmarks. However, their CNN outperformed the whole ROI extraction method proposed in [28] only when the Gaussian noise was added. Moreover, the evaluation was performed on images from CASIA-Palmprint database which have a black background.

Even though the previous works [13], [20], [39], [21], [22] already applied CNNs, their palmprint recognition pipelines were not trained end-to-end because the ROI was extracted before training [46], by some traditional methods [20], [22], [39] or even manually as in [21]. Moreover, in previous works, only linear transformations were considered in the ROI extraction step. In this paper, the proposed algorithm (see Section IV) considers non-linear deformation in the alignment and is end-to-end trainable, such that the power of deep learning and uncontrolled and uncooperative environment can be explored and addressed.

### B. Contactless Palmprint Databases

For contactless palmprint recognition, CASIA [47], IITD [48] and PolyU [49] databases are the most widely used benchmarks. Fig. 3 shows sample images from these three databases. CASIA-Palmprint database consists of 5502 grayscale hand images from 618 palms and is the largest publicly available contactless palmprint database in terms of the number of palms until now. The images were captured during one session and contain the palmar side of the hand with partial fingers on a black background. IITD Touchless Palmprint database provides 2601 colour hand images from 460 palms and also the corresponding ROI images. The images were taken during one session on the same background and have good illumination quality. PolyU Contactless 2D Palmprint database was collected from 177 hands. Each hand has ten colour images, taken during two sessions (5 images/session), under various illumination conditions with a black background. CASIA, IITD and PolyU databases are more suitable to study contactless palmprint recognition for commercial and governmental applications, rather than uncontrolled and uncooperative or forensic applications because the diversity of the images is not very large; there was some control over image acquisition parameters and high degree of user cooperation and the images were taken with an explicit goal of personal

TABLE I
The Details and Comparison of Benchmark Contactless Palmprint Databases and NTU Palmprint Databases.

| | NTU-PI-v1 | NTU-CP-v1 | IITD | PolyU | CASIA |
|---|---|---|---|---|---|
| No of palms | <u>2035</u> | 655 | 460 | 177 | 618 |
| No of images | <u>7781</u> | 2478 | 2601 | 1770 | 5502 |
| No of sessions | **NA** | 2 | 1 | 2 | 1 |
| Color image | Yes | Yes | Yes | Yes | No |
| ROI provided | No | No | Yes | Yes | No |
| Image size (pixels) | Mdn. 115x115 | Mdn. 1373x1373 | 1600x 1200 | 640x480 | 640x460 |
| ROI image size (pixels) | NA | NA | 150x150 | 128x128 | NA |
| All fingers visible | **Usually** | Yes | Yes | Yes | **No** |
| Space between fingers* | **Sometimes** | Usually | Yes | Yes | Yes |
| Posing variations* | **Large** | Small | Small | Small | Small |
| Affine transformations* | **Large** | **Medium** | Small | Small | Small |
| Complex background* | **Yes** | No (bright) | No (dark) | No (black) | No (black) |
| Illumination variations* | **Large** | **Medium** | Small | **Medium** | Small |
| Taken by one camera | **No** | **No** | Yes | Yes | Yes |
| Taken for palmprint recognition | **No** | Yes | Yes | Yes | Yes |
| Hand landmarks provided | Yes | Yes | No | No | No |

*Based on our visual assessment (see Figs. 3 and 4). The indicators of the uncontrolled and uncooperative environment are highlighted.

identification (see details in Table I). Thus, a new database is needed for this study.

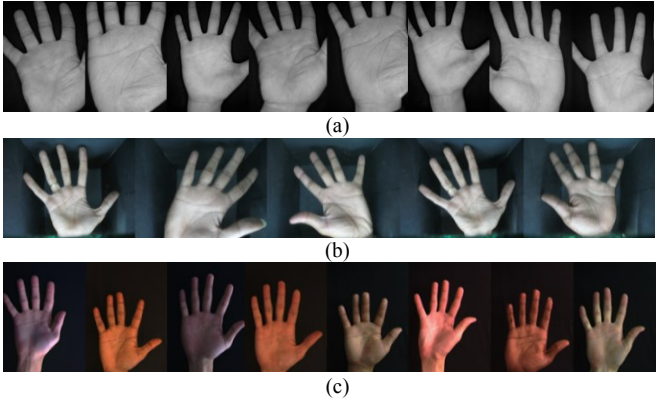

Fig. 3. Examples of images from (a) CASIA, (b) IITD, (c) PolyU contactless palmprint databases.

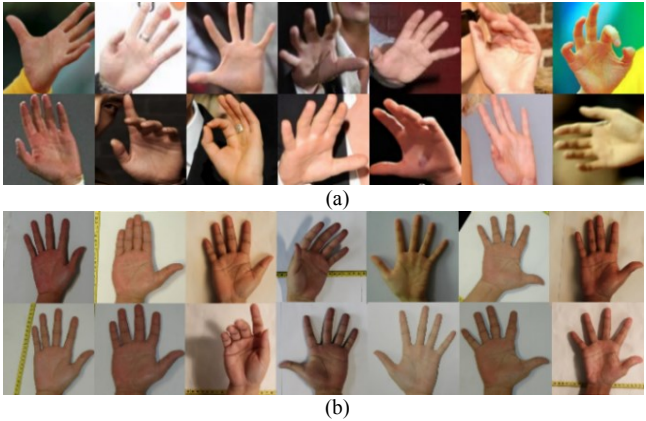

Fig. 4. Sample images from (a) NTU-PI-v1 and (b) NTU-CP-v1 databases. The images in the same column in (a) and (b) are from the same palm. Images are resized to the same size.

## III. NTU Palmprint Databases

For studying the uncontrolled and uncooperative palmprint recognition, the NTU Palmprints from the Internet (NTU-PI-v1) database, which consists of 7781 hand images collected from 2035 different palms of 1093 subjects with different ethnicity, sex and age, is established. According to our best knowledge, it is the largest publicly available palmprint database in terms of both palms and subjects. The images were collected and downloaded from the Internet galleries (see Fig. 5); then hand regions were manually cropped using a square bounding box. It is the same practice as in IARPA Janus Benchmark-A, one of the unconstrained face recognition benchmark [50], to avoid a bias of the database towards a "detector". Note that hand detection is out of scope in this study and one can employ F-RCNN or other detection methods for this task. There is no publicly available palmprint database for the target forensic application. Thus, the uncontrolled and uncooperative images downloaded from the Internet can be considered as a substitution to simulate the forensic environment. The database aims to reflect palm's diversity in the scenarios, where there is no control over image acquisition parameters and no subject's cooperation and the images are taken without any intention of palmprint recognition. The diversity is represented by significant differences in hand gestures, viewpoint, illumination, background, image quality and resolution (see Figs. 4a and 5). The hand image sizes vary from 30 by 30 pixels to 1415 by 1415 pixels and the median size is 115 by 115 pixels. Note that the sizes represent full hand images with fingers, not only palmprint ROI, which is likely to be two times smaller. In addition to the images, this database also provides annotated hand landmarks (see. Section IV-A) and segmented hand images.

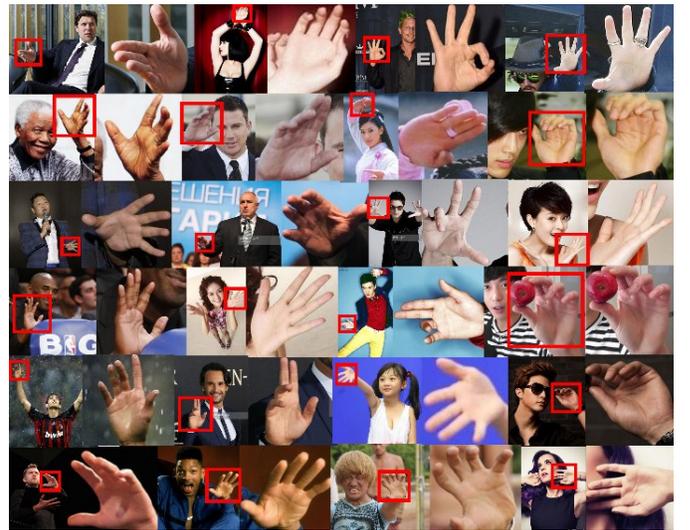

Fig. 5. Examples of original images downloaded from the Internet. Hands are highlighted in the original images.

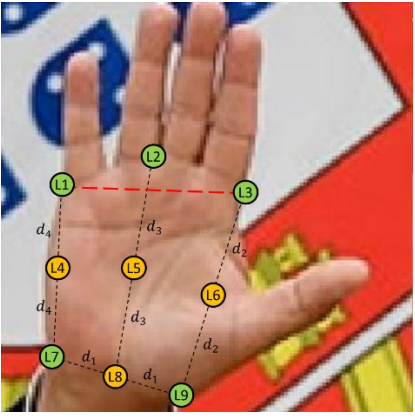

Fig. 6. The proposed initial configuration of hand landmarks.

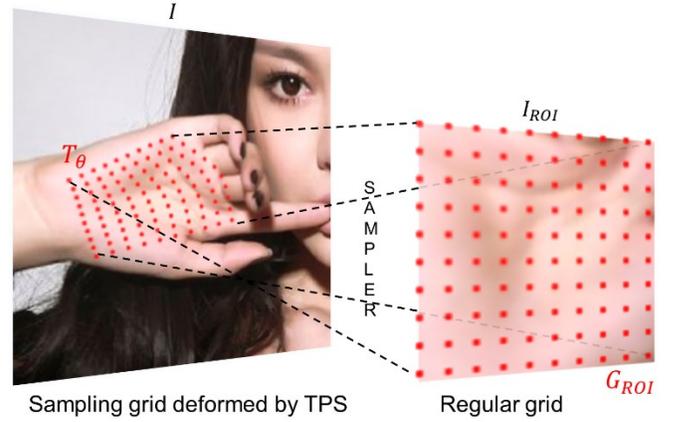

Fig. 7. Palmprint ROI extraction using TPS and a sampler. Sampling grid $T_\theta$ is created by TPS based on the hand landmarks in a hand image $I$. The sampler takes $T_\theta$ and $I$, and returns palmprint ROI $I_{ROI}$ by sampling $I$ into a regular grid $G_{ROI}$. The red dots are a schematic visualization of the sampling gird.

Another new palmprint database is NTU Contactless Palmprint (NTU-CP-v1) database, which contains 2478 images from 655 palms of 328 subjects, mainly Asian – Chinese, Indian, Malay, and some Caucasian and Eurasian persons. The database was collected in Singapore during two sessions without strict pose requirements in a contactless environment. The images were taken by Canon EOS 500D or NIKON D70s cameras and the hand regions were cropped. The hand image sizes vary from 420 by 420 pixels to 1977 by 1977 pixels and the median size is 1373 by 1373 pixels. Some sample images from the two NTU databases are shown in Figs. 4 and 5. The details and comparison of the palmprint databases are given in Table I. NTU-PI-v1 and NTU-CP-v1 will be available online [51] for research purpose in three months after this paper is published.

## IV. THE ARCHITECTURE DESIGN

In this section, the design of the network architecture is described. At the beginning, the motivations for particular choices in the proposed architecture such as the proposed hand landmarks initialization and configuration (Section IV-A), type of transformation (Section IV-B), and a CNN for feature extraction (Section IV-C) are discussed. Then, the details of the proposed, end-to-end architecture are given (Section IV-D).

### A. Hand Landmarks

Usually, in conventional palmprint recognition methods (see Section II), the ROI is defined based on the landmarks on the extracted hand contour, in particular in-between fingers. In the controlled environments, these landmarks are well-defined, stable, and allow to handle affine deformation of hand images because the hand contour can be extracted easily without error. However, such ROI extraction is sensitive to 3D pose variations, elastic palm deformations, unclear hand contours and complex background, which always appear in the uncontrolled environment. In other words, the traditional methods based on thresholding, edge detectors and boundary tracking work well only if the background and the hand have a significant color difference and there are clear spaces between fingers. Nonetheless, in the uncontrolled environment, such assumptions do not hold and the traditional methods will most likely fail in the detection. Thus, a CNN for hand landmark detection (see Section IV-D and Fig. 10: Localization Network), additional landmarks and non-affine transformation (see Section IV-B) are considered in this study.

In this study, landmarks are used to correct elastic deformation caused by different hand poses. Hence, the landmarks must be more than three in order to be able to parametrize non-rigid transformations. Note that three points can only determine affine transformations. Furthermore, the landmarks have to span the entire palm because more skin area carries more discriminative information [8]. Hence, the initial landmarks are defined at the four corners of the palm, which are labeled as L1, L3, L7 and L9 in Fig. 6. The landmarks have their corresponding points in a template. If there are only four landmarks, spanning the palm area, the line between L1 and L3 will cut across the palm and some palm information will be lost and the clear landmark L2 will be ignored (see red dashed line in Fig. 6). However, if L2 is used as a landmark, the two issues do not hold anymore and transformations with higher degrees of freedom can be defined for more accurate alignment. L1-L3 are well-defined, situated at the bottom of the fingers, whereas the exact locations of L7 and L9 are not always very clear. L7 and L9 are placed approximately at the border between the palm and wrist. In addition to these landmarks, the middle points of the lines, {L1, L7}, {L7, L9} and {L3 L9} are automatically extracted and labelled respectively as L4, L8 and L6. The middle point between L8 and L2 is also extracted and labelled as L5. In this study, nine landmark configuration is proposed. Note that the proposed hand landmark configuration is used for the Localization Network training (see Section V-D).

### B. Spatial Transformer

The aim of palm ROI extraction is to align each palmprint into the same coordinate system. A module that enables efficient spatial image manipulation within a neural network is a Spatial Transformer [52]. Thus, in this work, the Spatial Transformer is used for the ROI extraction (see Section V-D). The module consists of a trainable Localization Network, which regresses transformation parameters $\theta$, a sampling grid generator and a sampler. It can be used to implement any transformation $T_\theta$ that is differentiable with respect to its parameters $\theta$, e.g., affine, plane projective or thin plate spline (TPS) transformations. The module is differentiable and can be put into any place of a CNN architecture forming a Spatial Transformer Network, which can be trained with standard backpropagation. Jaderberg et al. [52] showed that the Spatial Transformer Network with TPS [53] is the most powerful for elastically deformed digits. Also, some recent state-of-the-art feature matching and alignment methods use Spatial Transformer Network with TPS [54], [55], [56]. The details of TPS can be found in [53]. Another reason for using the TPS transformation is that it is non-rigid and can be parametrized by control points in the Cartesian coordinate system, which are $x, y$ coordinates of hand landmarks in this study. Thus, in this work, hand landmark coordinates $\theta$ are used to parametrize the TPS deformation of the sampling grid. The illustration of the TPS palm transformation based on the proposed hand landmarks is shown in Fig. 7.

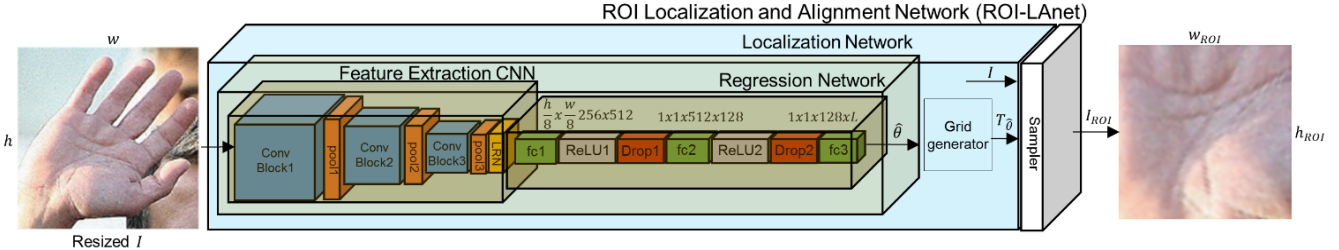

Fig. 8. The ROI-LAnet architecture. The input is the hand image $I$ resized to $h \times w$ pixels and the output is palmprint ROI image $I_{ROI}$.

### C. A Building Block for Palmprint Alignment and Feature Extraction

The convolutional layers from the VGG-16 network are used as a basic building block for palmprint alignment and feature extraction. The details of the VGG-16 can be found in [57]. The proposed alignment scheme does not use the entire VGG-16 network. The VGG-16 is first pre-trained on ImageNet and then the top layers are pruned. The rationale and motivation of such choice are that the palmprint databases are relatively small and training deep networks from scratch is very sensitive to overfitting. However, using a pre-trained network and reducing its depth can alleviate the data demand. Furthermore, palmprint consists of lines and texture. These lower level features also appear in natural images and can be extracted by the first several layers of a network trained on natural images [15]. On the other hand, features from the top layers carry more semantic meanings and therefore, the top layers trained on natural images may not accurately represent human hands. The modified pre-trained VGG is used as a building block (e.g., see Fig. 8: Feature Extraction CNN) in the Localization and Feature Extraction and Recognition Networks (see Fig. 10: Stage II). Note that these two building blocks in the proposed algorithm do not share parameters, because one operates on full hand images, which contain fingers and background whereas the other one operates on ROIs, which contain palmprint only.

### D. The Proposed Architecture

The proposed End-to-End Palmprint Recognition Network (EE-PRnet) consists of two main networks, ROI Localization and Alignment Network (ROI-LAnet) and Feature Extraction and Recognition Network (FERnet). Figs. 8 and 10 illustrate the architecture of the ROI-LAnet and EE-PRnet, respectively.

To align all palmprints into the same coordinate system and localize their ROIs, ROI-LAnet takes an original hand image $I$ resized to $h \times w$ pixels as an input and outputs palmprint ROI image $I_{ROI}$. The first part of the ROI-LAnet is the modified VGG-16 discussed in Section IV-C. More precisely, the pre-trained VGG-16 network [57] which is pruned after the layer $pool3$ with local response normalization (LRN) on top and is used as a feature extractor in the ROI-LAnet. This setting produces L2-normalized feature maps $f_h$, which retain the spatial information in the hand image. The feature maps $f_h$ are connected to the second part of the ROI-LAnet, which is a fully connected Regression Network with two hidden layers $fc1$ and $fc2$ with 512 and 128 neurons, respectively. Both layers are followed by ReLU activations and dropout to avoid neurons co-adaptation [58] and serve as a ROI augmentation mechanism in the training (see Section V-C). The Feature Extraction Network together with the Regression Network form the Localization Network (see Figs. 8 and 10), which outputs normalized coordinates $\hat{\theta}$ of the hand landmarks (see Section IV-A). The normalization range is between -1 and 1 and the normalized coordinates $\hat{\theta}$ are forwarded to the grid generator which transforms a regular, square grid $G$ to a deformed grid $T_{\hat{\theta}}(G)$ based on $\hat{\theta}$. A bilinear sampler takes the deformed grid $T_{\hat{\theta}}(G)$ as an input and samples the original hand image $I$ (not the resized $I$ inputted to the ROI-LAnet) to form a regular grid of $h_{ROI} \times w_{ROI}$ pixels, which is the ROI image, $I_{ROI}$ (see Fig. 8). Note that because of the coordinate normalization, the sampler can take an image of any size.

The ROI-LAnet is connected to the Feature Extraction and Recognition Network (FERnet) (see Fig. 10), which is responsible for palmprint feature extraction and recognition. To extract palmprint features, the ROI image $I_{ROI}$, is passed to another independent CNN, which is also a VGG-16 pruned after $pool3$ layer with the LRN on top. Even though its structure and the structure of the feature extraction network in the ROI-LAnet are same, there is no weight sharing between them. The output from the LRN layer, $f_{sROI}$ is a $h_{ROI}/8 \times w_{ROI}/8 \times 256$ spatial representation of the palmprint ROI. Note that the spatial dimensions $h_{ROI} \times w_{ROI}$ are defined in the grid generator in the ROI-LAnet. Then, the LRN is connected to a dropout layer and an embedding layer *fc4*, which outputs the final 512 dimensional palmprint descriptor $f_{PD}$. The vector $f_{PD}$ is passed to another dropout layer and then fully connected layer *fc5*, which returns the palmprint labels *PL*. In the training, L2 loss is used to train the Localization Network in the ROI-LAnet before connecting it to FERnet and Softmax loss is used to train the FERnet and the EE-PRnet (see Sections V-C and V-D).

## V. EXPERIMENTS

In this section, the evaluation protocols, implementation details of the proposed algorithm and other recognition methods, training strategy and experiments are given.

### A. Evaluation Protocols

In all experiments, the databases are divided into training and testing sets. Only the images in the training set are used to train the networks. No images nor ROIs from any database are discarded from the evaluation. To increase the number of possible comparisons, the left hands are flipped into the right ones and considered as different subjects. The evaluation protocol for the CASIA and IITD databases is the same as in the recent palmprint survey [13], where the first four gallery images are for training and the rest for testing. The evaluation protocol for the PolyU and NTU-CP-v1 databases is the same as in [34].

The NTU Palmprints from the Internet (NTU-PI-v1) database is divided into training/gallery and testing/probe sets as follows: if there is an even number of images from the same palm, then the images are randomly split into 50% for training and 50% for testing; if there is an odd number of images from the same palm, then the one more image goes to the training set. In this split, each palm in the testing set has a corresponding palm in the training set and in the case of odd number of samples, the training set has one more sample. Splitting NTU-PI-v1 database in this manner results in 3380 images from 1805 palms for testing and 4501 images from 2035 palms for training. No outside palmprint is used for training. The images in the training set which have no corresponding images in the testing set can be considered as distractors, which make the recognition more difficult. As with the protocol in IARPA Janus Benchmark-A [50], which was designed to study unconstrained face recognition for a particular group of subjects, e.g., identification of terrorists or gangsters in a watch list, this protocol allows performing training on the gallery set. Cumulative match characteristic (CMC) curve is used as an evaluation metric, which is common for forensic applications [4], [8], [50], [24]. In addition, rank-1 and rank-30 identification accuracies are also used as evaluation metrics.

### B. Implementation Details

The proposed architecture is implemented in MATLAB using MatConvNet library [59] and the code will be publicly available [51]. The pre-trained convolutional blocks from the VGG-16 are used (see Section IV-C). The proposed networks are trained with ADAM optimizer [60] with momentum 0.9, learning rate 0.001, no weight decay and batch size 128. The average R, G and B values are subtracted from each channel in input images $I$. In the ROI-LAnet, the input images $I$ are resized to 56 by 56 pixels and thus the size of the feature $f_h$ is $7 \times 7 \times 256$. The ROI image $I_{ROI}$ size is set to $112 \times 112$ pixels and thus the $f_{sROI}$ dimension is $14 \times 14 \times 256$. The dropouts *Drop1*, *Drop2*, *Drop3* and *Drop4* are set to 0.2, 0.1, 0.5 and 0.5, respectively. The ReLU units in the Regression Network are leaky ReLU activations with the leak value of 0.1. In the newly initialized layers *fc1-fc5*, the weights and biases are randomly sampled from a Gaussian distribution with 0 mean and 0.001 variance. The number of transformation parameters $L = 18$ because there are nine landmarks (see Section IV-A) and each is defined by $x, y$ coordinates. The output size $Nclass$ of the last *fc5* layer is the number of different palmprints in a training set. For recognition, the softmax, one-against-all partial least square regression (PLS) [61], linear SVM and k-nearest neighbor (k-NN) classifiers are compared in Section V-E. When using PLS and SVM, the palmprint descriptors $f_{PD}$ extracted from EE-PRnet are standardized (z-score) as input features. If the features are from the target palmprint of the PLS or SVM, its label is assigned to 1, otherwise -1. The number of latent components in PLS is 25 in Sections V-E and V-F and 50 in the final setting in Section V-G. When using the k-NN, k=1 and the cosine distance is used. The training is run on a PC with a single GPU card GeForce GTX TITAN X.

Other palmprint recognition methods are also used in the evaluation. CompCode, OrdinalCode and DoN are run using the programs provided by the authors in [36]. The parameters in these methods are the same as in [36] for the CASIA, IITD and PolyU 2D databases and for NTU-PI-v1 and NTU-CP-v1 are the same as those used for PolyU 2D contactless database. In the DOC [35], CR-CompCode [37], LLDP [33] and HOL [34], the codes provided by their authors and the default parameters are used. LLDP and HOL are used with Gabor filters. In the HOL, kernel spectral regression discriminant analysis (KSRDA) [62] is used for feature dimension reduction. In the DSCN [39], the deep scattering features are extracted using the implementation provided in [63], and for the recognition, one-against-all linear SVM is used. In the MF-LBP [26], the uniform LBP [64] with radius 2 and 8 neighbours and $\chi^2$ distance are used. In the SIFT-IRANSAC [25], the parameters of the Gabor filter are $\omega = 5$, $\sigma = 7.25$ and $F = 0.03$ and the SIFT descriptors are detected using the VL_FEAT library [65] with the threshold set to 100 and matched with the threshold 1.6. In the PalmNet [22] and PCANet [42], the publicly available codes are used with the default parameters. In the PCANet, the weighted PCA (WPCA) is used for feature dimension reduction, which is the same setting as in the original paper [42], when the algorithm was applied to face recognition. In the method from [21], in the comparison named AlexNet-S, AlexNet is fine-tuned and the sum rule of SVM and Softmax scores are used for recognition.

Three state-of-the-art deep learning architectures VGG-16 [57], ResNet-50 [66] and GoogLeNet [67] are trained for the comparison. All these networks are first pre-trained on the ImageNet dataset and then further fine-tuned on a palmprint database with ADAM optimizer, momentum 0.9, no weight decay and batch size 128, until convergence. The ROI images are resized to the default $224 \times 224$ pixels. For all these networks, the learning rate 0.001 is applied to the classification (last) layers only, whereas the lower layers have 100, 10 and 10 times smaller learning rates for VGG-16, ResNet-50 and GoogLeNet, respectively. VGG-16 and GoogLeNet are fine-tuned for 20 epochs, whereas the convergence of ResNet-50, which is very deep, takes 60 epochs.

### C. Training Strategy

In the experiments, different training and data augmentation strategies are investigated to analyze the components of the proposed networks. As the starting point for all the training stages, not to destroy the pre-trained convolutional layers, only the randomly initialized top layers are trained. Then, when the top layers are well initialized, lower layers can also be trained. In Stage I, the Localization Network is trained (Section V-D) to initialize the ROI-LAnet, which is used to extract palmprint ROIs. In Stage II, the FERnet is trained based on palmprint ROIs. In Stage III, to investigate the end-to-end training of the EE-PRnet (based on full hand images as inputs), the initialized ROI-LAnet from Stage I is connected to the FERnet and trained together. In other words, Stage III is an end-to-end training, whereas Stage II is not. Note that the FERnet in Stage III is not the trained FERnet from Stage II.

In Stage II, the input is the palmprint ROI, whereas in Stage III, the full hand image. When training the EE-PRnet, the palmprint ROI inputted to the FERnet depends on the input hand image $I$ and the ROI-LAnet. The results from Stage II and Stage III are expected to be different because 1) the dropouts *Drop1*, *Drop2*; 2) the parameters updates in the Localization Network in the ROI-LAnet and 3) on-the-fly data augmentation of hand images $I$, which will allow that in the same hand image $I$, the different hand landmarks $\hat{\theta}$ can be estimated. Thus, different palmprint ROIs $I_{ROI}$ can be extracted in training. In other words, the ROI-LAnet performs on-the-fly, in-network ROI augmentation, whose contribution is analyzed in experiments in Section V-E. Fig. 9 illustrates the ROI augmentation based on the dropouts in the Localization Network.

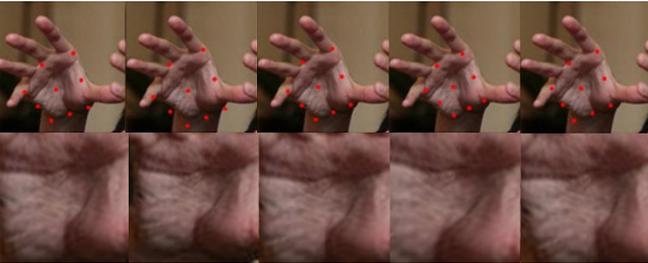

Fig. 9. Illustration of the in-network ROI augmentation. Top row: examples of different hand landmarks $\hat{\theta}$ (red dots) detected by the Localization Network in the same hand image $I$. Bottom row, the same column: the corresponding palmprint ROI images $I_{ROI}$, returned by ROI-LAnet. Zoom in recommended for the best view.

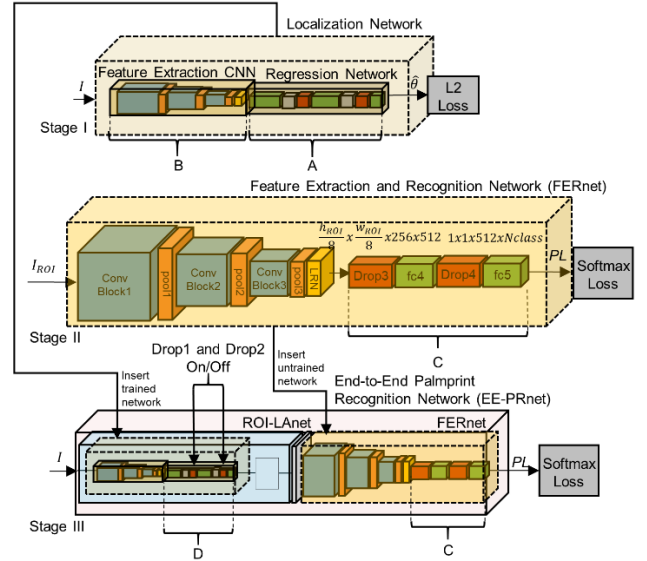

Fig. 10. The proposed network architectures and training schemes.

### D. Localization Network Training

The Localization Network is trained on annotated hand images from the NTU-PI-v1 (see Section III and IV-A) with the L2 loss (see Fig. 10). The network is fed with the original and segmented hand images from the training set. For data augmentation, three hand images rotated by 90, 180 and 270 degrees and three randomly rotated segmented images are used. At first, to train the Regression Network and not to destroy the pre-trained convolutional blocks, only the parameters from layers *fc1, fc3* denoted as block A (Fig. 10: Stage I-A) are being trained for 10 epochs. Then, in the next sequence, when the Regression Network is well-initialized, the convolutional layers denoted as block B are unfrozen and trained together (with block A) for the next 15 epochs (Fig. 10: Stage I-AB). Such pre-trained Localization Network is inserted into the ROI-LAnet and used for the ROI extraction.

The accuracy of ROI extraction is measured by normalized mean error (NME) [68] of the detected hand landmarks, which is $1.99\pm0.36\%$ after the Localization Network training. It should be emphasized that the objective is palmprint recognition and the Localization Network training is for initializing the network with reasonably good weights before the EE-PRnet training. Moreover, some landmarks such as L7 and L9 are not well defined and their ground truth annotations are expected to have higher errors than other landmarks. Thus, the Localization Network is further trained in the EE-PRnet with Softmax loss.

### E. Experiments on NTU Palmprints from the Internet

This subsection aims to analyze the components of the proposed EE-PRnet by employing different training strategies $S$ on the NTU-PI-v1 and four different classifiers for recognition. The strategies' details are given in Table II. In strategies *S1-S5*, the ROI-LAnet and the FERnet are connected and the EE-PRnet is trained based on hand images, whereas in *S0*, only FERnet is trained separately based on ROI images. In the beginning, for all the strategies, only the top layers *fc4* and *fc5* denoted as block C (see Fig. 10) are trained. Then, when the network converges which typically occurs after 20 epochs, the Localization Network can also be fine-tuned and the training is stopped at epoch $N$=40. For the color data augmentation denoted as $ct$, the saturation and contrast of each image are randomly shifted on-the-fly.

TABLE II
DETAILS OF THE DIFFERENT TRAINING STRATEGIES ($S$).

| | | FERnet (fc4, fc5) | | ROI-LAnet (fc1, fc2, fc3) | | (Drop1, Drop2) | |
|---|---|---|---|---|---|---|---|
| $S$ | Stage | tune | epoch | tune | epoch | switch | epoch |
| $S0$ | II | C | $[1,N]$ | - | - | - | - |
| $S1$ | III | C | $[1,N]$ | × | $[1,N]$ | off | $[1,N]$ |
| $S2$ | III | C | $[1,N]$ | × | $[1,N]$ | on | $[1,N]$ |
| $S3$ | III | C | $[1,N]$ | D | $(20,N]$ | off | $[1,N]$ |
| $S4$ | III | C | $[1,N]$ | D | $(20,N]$ | on | $[1,N]$ |
| $S5$ | III | C | $[1,N]$ | D | $(20,N]$ | on/off | $[1,35]/(35,N]$ |

In the first and most basic strategy $S0$, the ROI images are extracted by the ROI-LAnet and input to train the FERnet (Stage II-C). Note that strategy $S0$ is not end-to-end. During the training using the strategy $S1$, all the layers in the ROI-LAnet and the *ConvBlock1-3* in the FERnet are frozen and only the parameters in block C are being updated for $N$ epochs, while the dropouts $Drop1$, $Drop2$ are switched off (Stage III-C-off). The strategy $S2$ is the same as $S1$, but the dropouts $Drop1$, $Drop2$ are switched on (Stage III-C-on). The strategies $S3$ and $S4$ are the same as the $S2$, but the block D (see Fig. 10) in the ROI-LAnet is fine-tuned after epoch 20 with the learning rate 0.0001 and the dropouts $Drop1$, $Drop2$ are switched off for the $S3$ (Stage III-CD-off) and on for the $S4$ (Stage III-CD-on). Finally, the strategy $S5$ is the same as $S4$ but after epoch 35 (from Stage III-CD-on) the dropouts $Drop1$, $Drop2$ are switched off (Stage III-CD-on/off). Additionally, the results, when no color augmentation $ct$ is used in the strategy $S0$, are also provided ($S0nct$). To show the merit of the ROI extraction by the ROI-LAnet, hand images are input directly to the FERnet (Stage II-C), and the results are reported as $S0h$. The comparison of the proposed training strategies and four classifiers is presented in Table III.

The results show that connecting two networks ROI-LAnet and FERnet into EE-PRnet, benefits the overall performance (see $S0$ vs $S1$) and PLS outperforms Softmax, SVM and k-NN. The dropout based augmentation by the ROI-LAnet also positively contribute ($S1$ vs $S2$) and its significance is even more noticeable when the Regression Network (block D) is

TABLE III
RANK-1 AND RANK-30 ACCURACY (%) OF DIFFERENT TRAINING STRATEGIES ($S$) AND CLASSIFIERS ON NTU PALMPRINTS FROM THE INTERNET DATABASE.

| | PLS | | Softmax | | SVM | | k-NN | |
|---|---|---|---|---|---|---|---|---|
| $S$ | Rank-1 | Rank-30 | Rank-1 | Rank-30 | Rank-1 | Rank-30 | Rank-1 | Rank-30 |
| $S0$ | 37.69 | 60.53 | 24.05 | 50.05 | 28.67 | 52.69 | 18.49 | 43.37 |
| $S1$ | 39.32 | 62.04 | 24.94 | 52.04 | 28.52 | 53.52 | 19.70 | 44.62 |
| $S2$ | *39.82* | *63.52* | 28.14 | 56.33 | 33.64 | **59.14** | 24.35 | 51.45 |
| $S3$ | 38.96 | 62.25 | 24.97 | 51.86 | 28.76 | 53.43 | 19.44 | 44.88 |
| $S4$ | **40.80** | 63.70 | 29.05 | *57.04* | 33.67 | *59.02* | **24.91** | **51.57** |
| $S5$ | **40.80** | **64.20** | **29.70** | **57.96** | **33.85** | 58.96 | 24.23 | 51.36 |
| $S0nct$ | 37.07 | 60.18 | 23.46 | 49.08 | 26.12 | 51.80 | 17.31 | 42.16 |
| $S0h$ | 14.56 | 32.19 | 10.12 | 30.18 | 10.30 | 27.57 | 8.22 | 26.54 |

The **first**, second and third best strategies are highlighted.

TABLE IV
RANK-1 AND RANK-30 ACCURACY (%) OF THE SELECTED TRAINING STRATEGIES ($S$) ON IITD, CASIA AND POLYU DATABASES.

| | IITD | | PolyU | | CASIA | |
|---|---|---|---|---|---|---|
| $S$ | Rank-1 | Rank-30 | Rank-1 | Rank-30 | Rank-1 | Rank-30 |
| $S0$ | 96.19 | 99.08 | 98.98 | 99.55 | - | - |
| $S2$ | 97.11 | 99.08 | 99.55 | 99.89 | 94.64 (94.14)* | 97.78 (97.55)* |
| $S4$ | **98.69** | 99.47 | 99.66 | **99.89** | 96.82 (94.48)* | 99.40 (98.21)* |
| $S5$ | 98.16 | **99.61** | **99.66** | **99.89** | **97.52 (94.91)*** | **99.54 (98.31)*** |

*The numbers in the brackets for the CASIA database are the results without using data augmentation $at$.

fine-tuned ($S3$ vs $S4$). The dropouts' decay for the last 5 epochs while fine-tuning also can improve the performance ($S4$ vs $S5$). In summary, when training on NTU-PI-v1 images with PLS as a classifier, the proposed strategy with connected networks and end-to-end training outperforms the separated networks setting ($S0$ vs $S5$) with a margin of 3.11% and 3.67% at rank-1 and rank-30, respectively. Moreover, the ROI extraction by the ROI-LAnet is an essential step ($S0h$ vs $S0$) and without it, the performance drops by 23.13%.

### F. Experiments on CASIA, IITD, and PolyU Contactless Palmprint Benchmark Databases

The top three strategies $S2$, $S4$ and $S5$ from Section V-E are used to train the EE-PRnet for the CASIA, IITD and PolyU databases. To compare the ROI extraction performance, the strategy $S0$ is also applied when the databases, i.e., the IITD and PolyU databases provide extracted ROI. Because the proposed EE-PRnet requires square image input, some simple automatic pre-processing steps are applied on the CASIA, IITD and PolyU images. Note that the network architecture, training and testing policies are fixed and exactly the same as described in Section V-E.

The images in the CASIA database, which have a black background are padded with zeros to make them square. In the IITD database, only the fixed central part is cropped. In the PolyU database, the images except for hands also contain a significant portion of a black background. To detect the hands only, Otsu thresholding and the square bounding box around the biggest connected component are used.

Such pre-processed images are used to train the EE-PRnet for $N$=40 epochs. In strategies $S4$ and $S5$, the learning rate in block D is set to 0.001, which is 10 times higher than in Section V-E, because the block D is a part of the Localization Network, which is pre-trained on the NTU-PI-v1 (Section V-D). To compare the FERnet trained on ROIs extracted by traditional methods with the EE-PRnet, the ROIs provided in the IITD and PolyU databases are used to train the FERnet with the strategy $S0$. The results are presented in Table IV. The results show that even though the Localization Network is pre-trained on the different database (NTU-PI-v1), it can generalize well and can be fine-tuned to a given database ($S2$ vs $S4/S5$).

However, we observe slightly lower performance on the CASIA database (see the CASIA results in the brackets in Table IV) which can be caused by the discrepancy between the CASIA and NTU-PI-v1 (see Figs. 3 and 4a and Table I). Contrary to NTU-PI-v1, the hand images in the CASIA

TABLE V
Rank-Ordered Accuracy (%) and EER (%) of Different Palmprint Recognition Methods on Five Databases.

| | NTU-PI-v1 | | NTU-CP-v1 | | IITD | | PolyU | | CASIA | |
|---|---|---|---|---|---|---|---|---|---|---|
| | Rank-1 | Rank-30 | Rank-1 | EER | Rank-1 | EER | Rank-1 | EER | Rank-1 | EER |
| Proposed | **41.92** | **64.73** | **95.34** | **0.76** | **99.61** | **0.26** | 99.77 | **0.15** | **97.65** | **0.73** |
| CompCode | 5.41 | 15.00 | 23.62 | 30.92 | 77.79*/97.76 | 1.39 | 99.21** | 0.68** | 79.27*/97.32 | 1.08 |
| OrdinalCode | 29.34 | 49.94 | 50.59 | 19.10 | 73.26*/97.50 | 2.09 | 99.55** | 0.23 | 73.32*/96.32 | 1.75 |
| DoN | 25.47 | 46.30 | 52.62 | 19.51 | 98.16 | 1.45 | **100.00**** | 0.22** | 96.89 | 1.16 |
| DOC | 20.68 | 39.08 | 36.61 | 31.86 | 89.99*/90.01 | 16.79 | 90.96 | 8.31 | 78.51*/82.40 | 18.71 |
| DSCN | 15.47 | 33.70 | 49.15 | 11.95 | 97.10 | 0.51 | 82.25 | 6.21 | 95.23 | 0.79 |
| SIFT-IRANSAC | 10.14 | 21.24 | 74.75 | 15.26 | 91.46*/99.47 | 1.45 | 88.58 | 15.74 | 93.28*/96.69 | 3.46 |
| MF-LBP | 5.91 | 20.30 | 16.86 | 26.92 | 63.60 | 19.29 | 79.32 | 10.19 | 72.41 | 11.30 |
| ResNet-50 | 4.46 | 17.31 | 7.28 | 28.47 | 95.57*/76.74 | 3.68 | 54.92 | 13.45 | 95.21*/74.40 | 4.27 |
| GoogLeNet | 4.40 | 21.09 | 27.97 | 14.88 | 96.22*/87.91 | 1.45 | 68.59 | 9.96 | 93.84*/89.02 | 1.65 |
| VGG-16 | 8.79 | 27.54 | 28.05 | 13.14 | 92.12*/86.47 | 1.97 | 73.33 | 11.19 | 94.01*/89.02 | 1.65 |
| PalmNet | 20.21 | 36.21 | 68.47 | 16.95 | 98.42 | 1.31 | 99.77 | 0.24 | 97.02 | 1.92 |
| PCANet | 20.56 | 45.62 | 72.03 | 7.76 | 97.37 | 1.18 | 99.66 | 0.45 | 95.53 | 1.46 |
| AlexNet-S | 11.78 | 29.85 | 31.86 | 17.72 | 97.24 | 0.92 | 70.96 | 14.91 | 92.76 | 1.79 |
| CR-CompCode | 23.37 | 45.27 | 52.63 | 13.31 | 94.22 | 2.78 | 97.18 | 1.02 | 91.73 | 3.18 |
| LLDP | 9.44 | 24.47 | 51.61 | 18.19 | 95.17*/98.16 | 1.74 | 99.21 | 0.68 | 93.00*/95.73 | 2.67 |
| HOL | 6.33 | 11.42 | 64.75 | 10.67 | 95.93 | 0.66 | 99.44 | 0.42 | 95.20 | 1.32 |

The numbers with *, ** and without any star indicator are from the recent review [13], the DoN paper [36] and our experiments, respectively.

database are grayscale and usually, the full fingers and thumbs are not visible. The visual difference between the CASIA database and the other databases can be seen in Fig. 3. Thus, a different augmentation technique is applied to refine the training strategy for this database. Rather than using color data augmentation scheme *ct* on CASIA greyscale images taken under good illumination, a more aggressive spatial data augmentation scheme denoted as *at* is used because the images have significant scale and translation variations. In the augmentation scheme *at*, each input hand image is randomly either rotated (+/- 5, 20, 90 deg.) or scaled (0.8, 0.9, 1, 1.1, 1.2) or translated (+/- 0.1, 0.15, 0.2) on-the-fly. The results (see Table IV: CASIA) show that the proposed strategies combined with the augmentation scheme *at* benefit the performance on the CASIA which is visually quite distant from the target database NTU-PI-v1, for which the algorithm is designed. In the next subsection, the training strategy *S5* is combined with the augmentation scheme *at* for all the databases and the final results of the proposed algorithm are reported.

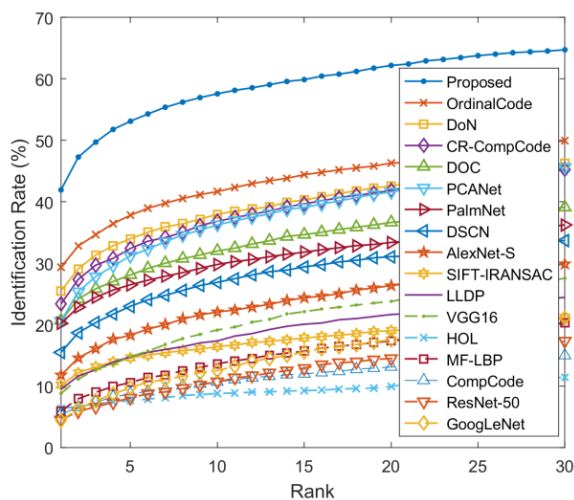

Fig. 11. CMC curves of the proposed algorithm and state-of-the-art palmprint recognition and deep learning methods on NTU-PI-v1 database.

### G. Comparison with the State-of-the-Art Palmprint Recognition Methods

In this section, the comparison of the proposed algorithm with the state-of-the-art palmprint recognition and deep learning methods on the representative contactless palmprint benchmarks (Section II-B) and the NTU-Palmprint databases (Section III) is given.

For each database, the proposed EE-PRnet is trained for $N$=60 epochs with the *S5* strategy. Data augmentation scheme *ct* is used for all the databases, except the CASIA, in which the scheme *at* is used. After 40 epochs the augmentation *at* (see Section V-F) is also applied to all databases. The evaluation protocols are described in Section V-A. In the experiments on NTU-PI-v1 and NTU-CP-v1, the ROIs extracted by the pre-trained ROI-LAnet (Section V-D) are input to CompCode [29], OrdinalCode [30], DOC [35], DoN [36], DSCN [39], SIFT-IRANSAC [25], MF-LBP [26], VGG-16 [57], ResNet-50 [66], GoogLeNet [67], PalmNet [22], PCANet [42], AlexNet-S [21], CR-CompCode [37], LLDP [33] and HOL [34] because their preprocessing steps cannot handle images in these two databases, which have complex backgrounds or no gaps between the fingers. In other words, if the proposed ROI-LAnet is not used to "help" other palmprint recognition methods to extract ROI, all of them would fail. The implementation details of these methods are described in Section V-B. The ROIs provided in the IITD and PolyU and the ROI extracted from CASIA databases by us using the method from [28] are used as inputs for all the methods, except for the proposed algorithm.

Table V compares the performance of the 17 methods on the 5 databases. To compare the methods on the CASIA and IITD databases, the numbers with * are taken from the most recent palmprint survey [13]. The difference between the numbers with * taken from [13] and the numbers from our experiments can be caused by different parameters and fine-tuning techniques. These parameters and fine-tuning settings are not provided in [13]. Thus theirs and our results are given. To compare the performance on the PolyU database, the numbers with ** are taken from [36]. The numbers without any star indicator are from our experiments. Note that some results from our experiments are obtained by using the codes or program provided by the authors (see Section V-B). Fig. 11 shows CMC curves for the NTU-PI-v1. On the NTU-PI-v1, the proposed algorithm achieves 41.92%, 59.88% and 64.73% at rank-1, rank-15 and rank-30, respectively. The results show that the proposed algorithm outperforms the existing methods on the NTU-PI-v1 with a margin of 12.58% and 14.79% at rank-1 and rank-30, respectively. To visualize the characteristic of the proposed algorithm performance, examples of top-10 retrievals on the NTU-PI-v1 database are shown in Fig. 13. The proposed algorithm also achieves the highest identification performance on the CASIA and IITD databases. In [13], the authors showed that the deep learning based methods can achieve higher performance than conventional palmprint recognition methods. More clearly, the highest rank-1 accuracy on the CASIA and IITD databases was achieved by extracting ROI with a conventional method and then fine-tuning the ResNet-50 and GoogLeNet, respectively. Nevertheless, the proposed algorithm outperforms the ResNet-50 and GoogLeNet results reported in [11] with a margin of 2.44% and 3.39%, respectively. The proposed algorithm is unable to achieve the rank-1 accuracy of 100% on the PolyU database as it is reported for DoN in [36]. It achieves 99.77% rank-1 accuracy, which means that only 2 images are not retrieved at rank-1. It should be emphasized that in the experiment, the training/testing policy and hyper-parameters of the proposed algorithm are fixed for all databases whereas in [36], besides a carefully designed palmprint descriptor, the authors also carefully tuned the algorithm parameters for a given database, especially the offset which defines the horizontal and vertical translations to improve the alignment between two images. In other words, each image from the testing set is being translated $n_T$ times and compared with one training image resulting in $n_T$ comparison scores for one testing image. However, in the proposed algorithm, there is no such post-processing.

Because the existing palmprint recognition methods and NTU-CP-v1, CASIA, IITD and PolyU contactless palmprint databases are designed for commercial applications, likely used for verification, rather than identification, the equal error rates (EER, the smaller, the better) which measure the verification performance are also reported in Table V. The results show that the proposed algorithm outperforms all considered palmprint recognition methods on all the four databases. It achieves 0.76%, 0.26%, 0.15% and 0.73% EER on NTU-CP-v1, IITD, PolyU and CASIA databases, respectively, which are 7.00%, 0.25%, 0.07% and 0.06% less than the corresponding second best.

The proposed algorithm is also evaluated on two other contactless palmprint databases, namely, Tongji contactless palmprint dataset [37] and REST hand database 2016 [69], which were designed for contactless palmprint identification. The evaluation protocols for the Tongji and REST are the same as in the original study [37] and NTU-PI-v1, respectively. The proposed algorithm achieves 98.63% and 93.34% rank-1 identification accuracy on the Tongji and REST databases, respectively. These results are comparable with the results reported in their corresponding studies. Moreover, we also compare with the results reported for very recently published discriminant direction binary palmprint descriptor (DDBPD), on IITD, CASIA and Tongji databases. The highest rank-1 accuracy achieved by DDBPD [38] on IITD, CASIA and Tongji are 96.4373±1.1079%, 96.4085±2.6860% and 98.7333%, respectively. The proposed algorithm achieves comparable rank-1 accuracy of 99.61%, 97.65% and 98.63% on IITD, CASIA and Tongji databases, respectively. However, we could not compare on other databases, including NTU-PI-v1, because the code for DDBPD is not publicly available yet.

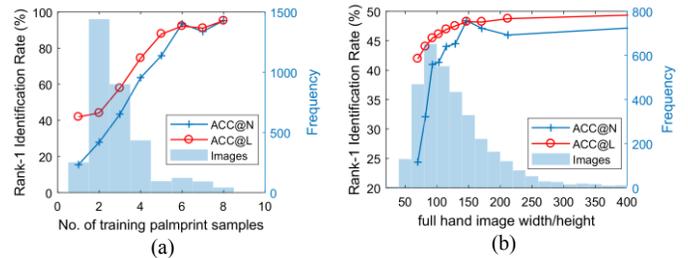

Fig. 12. The performance of the proposed algorithm as a function of (a) a number of available training samples per palmprint, (b) full hand image width/height in pixels.

### H. Performance Analysis

In this section, a performance analysis of the proposed algorithm on NTU-PI-v1 database is given. The comparison scores from Section V-G are used to analyze rank-1 identification accuracy ($ACC$) with respect to (a) a number of training palmprint samples and (b) the full hand image size. Thresholds of minimum and range for the number of training palmprint samples and image size are applied to calculate the respective accuracy. In other words, only the scores of probe images that satisfy the desired criteria are selected. Note that, the gallery size remains the same, which is 2035 palmprints. Fig. 12a/b shows $ACC@N(X)$, and $ACC@L(X)$, which are accuracy calculated from the rank-1 scores with $X$, and at least $X$ training palmprint samples/pixels in the probe image, respectively. For example (see Fig. 12a), if there are 4 training samples, $ACC@N(4)$=63.59; if there are at least 4 training samples, $ACC@L(4)$=74.43%.

On average, each palmprint has only 2.88 training samples in NTU-PI-v1 database, whereas PolyU, CASIA, and IITD databases have at least 4 gallery samples. In NTU-PI-v1 database, adding training samples almost linearly increases the performance to reach 95.24% for 8 samples. In forensics, the training samples and the gallery can be created from multiple sources, e.g., Internet images, videos, images taken by prison imagining systems, etc. The image size also affects the performance. However, it is less significant compared with the number of training samples. Note that in the first two bins of the histogram in Fig. 12b, the full image width/height ranges from 40 to 80 pixels meaning that the approximate palmprint ROI sizes have only 20×20 or 40×40 pixels. These show how challenging the NTU-PI-v1 database is. Nevertheless, to study forensic applications, the database has to be challenging and reflect the variety of different scenarios. More discussion about the accuracy in forensics is given in Section VI.

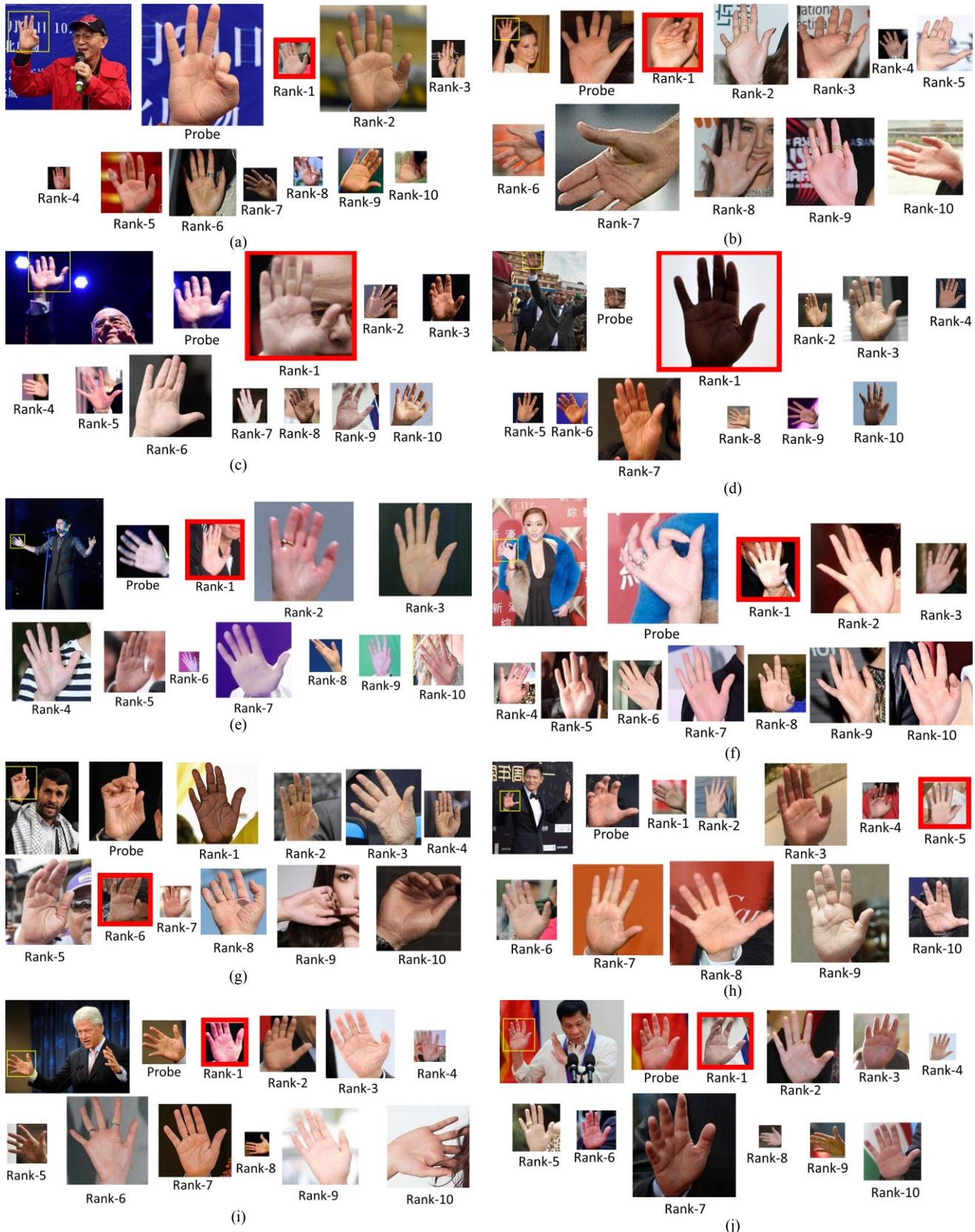

Fig. 13. Examples of top-10 identification results of the proposed algorithm on NTU-PI-v1 database. The probe images are in the yellow boxes in the original images. The true matches in the gallery are in the red boxes. In each subfigure, the relative scales of the hand images are preserved for comparison.

## VI. Conclusions and Discussion

In forensic investigation, criminal and victim identification based on digital images is very challenging if no obvious characteristics such as face, skin marks or tattoos are visible. Even though, in terrorist, riot or child sexual abuse images, criminals hide their faces, palms can be still visible, especially when the subjects raise their hands to salute, wave, cover the camera or touch the victim or offender. The existing palmprint recognition methods and databases were designed for 1) online palmprint recognition for commercial applications, which require a controlled environment and user cooperation, or 2) the latent palmprint identification for forensic applications, which require high-resolution latent prints collected from a crime scene. Contrary to the images in the existing palmprint studies, some evidence images are taken in uncontrolled and uncooperative environments and have no high-resolution features such as minutia or ridges. The merit of uncontrolled and uncooperative palmprint recognition for forensic investigation is not fully exposed yet. In this paper, this uncontrolled and uncooperative palmprint recognition problem is investigated.

The new NTU Palmprints from the Internet (NTU-PI-v1) database is provided to study uncontrolled and uncooperative palmprint recognition for forensic applications. The database aims to simulate forensic scenarios and contains 7781 images from 2035 different palms collected from the Internet, which makes it the largest publicly available palmprint database, in terms of palms and subjects. In addition to the NTU-PI-v1 database, the contactless palmprint database, NTU-CP-v1 is provided. In forensic, the database size is not always very large because the list of suspects can be narrowed down using hints and clues, such as gender, age, ethnicity, location, date, etc.

Most of the previous works use the feature-engineering approach and only a few use deep learning. Even though those employ deep learning, the whole pipeline is not end-to-end trainable because the ROI is extracted using some traditional methods and aligned using linear transformation, which are not applicable to the uncontrolled and uncooperative environment. In this paper, the end-to-end trainable EE-PRnet, which consists of the alignment network employing non-linear transformation, feature extraction network, and in-network ROI augmentation scheme based on dropout, is proposed. In the experiments, the network components of the proposed EE-PRnet and different training strategies are analyzed and their contributions to the performance are studied. The proposed EE-PRnet is compared with thirteen state-of-the-art palmprint recognition and three popular deep learning methods on five palmprint databases, including the uncontrolled and uncooperative NTU-PI-v1 and contactless NTU-CP-v1, CASIA, IITD and PolyU databases. The experimental results show that the proposed algorithm, which is designed for forensic applications, is also applicable to contactless palmprint recognition and outperforms the existing palmprint recognition methods. On the NTU-PI-v1 database, its performance is also significantly higher even though the other methods have the "help" from the proposed ROI-LAnet, which targets to extract ROIs in images collected in the uncontrolled and uncooperative environment.

The poor quality and uncontrolled images are common in forensic investigation, which makes the identification rates considerably lower than in cooperative and well-controlled environments. For latent palmprints in [24], Jain et al. showed in the CMC that the rank-1 accuracy is around 25% whereas rank-20 accuracy is slightly above 40%. Even well-studied fingerprint recognition methods applied to bad quality images achieve low accuracy, e.g., the rank-1 accuracy of 21.2-34.1% for commercial fingerprint matchers on fingerprint images with ugly quality reported by Paulino et al. [4]. In forensic investigation, not only rank-1 but also higher ranks, e.g., rank-30 are valuable because they can reduce the suspect list and the number of manual searches by forensic investigators. The proposed algorithm on the NTU-PI-v1 database achieved the rank-1 and rank-30 accuracy of 41.92% and 64.73%, respectively.

### Acknowledgment

The authors would like to thank Chinese Academy of Sciences, Institute of Automation, Hong Kong PolyU, IIT Delhi, Tongji University and University of Sfax, for sharing the palmprint databases.